\documentclass{article}

\usepackage[final]{corl_2024} 

\usepackage{amsmath,amssymb,amsfonts}
\usepackage{algorithmic}
\usepackage{graphicx}
\usepackage{textcomp}
\usepackage{xcolor}
\usepackage{cleveref}
\usepackage{subfig} 
\usepackage{subfloat}
\usepackage{fancyhdr}
\usepackage{enumitem}

\renewcommand{\vec}[1]{\boldsymbol{#1}}
\newcommand{\mat}[1]{\boldsymbol{\mathrm{#1}}}

\title{Estimating Neural Network Robustness via Lipschitz Constant and Architecture Sensitivity}

\author{
  Abulikemu Abuduweili\\
  Carnegie Mellon University\\
  \texttt{abulikea@andrew.cmu.edu} \\
  \And 
   Changliu Liu\\
  Carnegie Mellon University\\
  \texttt{cliu6@andrew.cmu.edu} \\
}

\begin{document}
\maketitle


\begin{abstract}
Ensuring neural network robustness is essential for the safe and reliable operation of robotic learning systems, especially in perception and decision-making tasks within real-world environments. This paper investigates the robustness of neural networks in perception systems, specifically examining their sensitivity to targeted, small-scale perturbations. We identify the Lipschitz constant as a key metric for quantifying and enhancing network robustness. We derive an analytical expression to compute the Lipschitz constant based on neural network architecture, providing a theoretical basis for estimating and improving robustness. Several experiments reveal the relationship between network design, the Lipschitz constant, and robustness, offering practical insights for developing safer, more robust robot learning systems.
\end{abstract}

\keywords{Robustness, Lipschitz Continuity, Robot Learning, Neural Networks} 


\section{Introduction}

Deep neural networks have been successfully applied to many tasks in robotics \cite{sunderhauf2018limits}, including perception \cite{voulodimos2018deep}, prediction \cite{abuduweili2019adaptable}, planning, and control \cite{grigorescu2020survey}.  
However, their sensitivity to input perturbations poses significant challenges \cite{szegedy2013intriguing}, particularly in safety-critical applications like autonomous driving and human-robot collaboration \cite{liu2023proactive}. For instance, a small but well-designed modification to an input image can cause a neural network used in perception systems to misclassify the image \cite{goodfellow2014explaining}, undermining its reliability in applications like autonomous driving or robotic navigation. Such issues raise concerns about the deployment of these models in real-world robotic systems, where safety and robustness are critical. 
In autonomous driving, adversarial examples could trigger unintended actions, such as incorrect path planning or object misdetection, potentially compromising the safety of autonomous systems.

In general, there are two different approaches one can evaluate the robustness of a neural network \cite{balunovic2020adversarial}: adversarial attack-based methods that demonstrate an upper bound, or verification-based methods that prove a lower bound.  Adversarial attack approaches are easy to conduct, but the upper bound may not be useful \cite{carlini2017towards}. Verification-based approaches, while sound, are substantially more difficult to implement in practice, and all attempts have required approximations \cite{liu2021algorithms}.
Besides the adversarial attack and verification-based methods for evaluating the robustness, some works study the neural network’s intrinsic robustness based on the Lipschitz continuity of neural network \cite{weng2018evaluating}. In deep neural networks, tight bounds on its Lipschitz constant can be extremely useful in a variety of applications: (1) The adversarial robustness of a neural network is closely related to its Lipschitz continuity \cite{latorre2020lipschitz}. Constraining local Lipschitz constants in neural networks is helpful in avoiding adversarial attacks \cite{weng2018evaluating}. (2)  Generalization bounds critically rely on the Lipschitz constant of the neural networks in deep learning theory\cite{bartlett2017spectrally}.  In these applications and many others, it is essential to estimate the Lipschitz constant both accurately and efficiently.

Various approaches have been proposed to measure and control the Lipschitz constant of neural networks \cite{fazlyab2019efficient}. Early work by Szegedy \textit{et al.} \cite{szegedy2013intriguing} introduced an upper bound based on spectral norms of linear layers.  Fazlyab \textit{et al.} developed a convex programming framework for tighter bounds, and  Shi \textit{et al.} computed relatively precise Lipschitz constants using bound propagation techniques\cite{shi2022efficiently}.  While these methods provide numerical estimates, they do not fully explore the relationship between neural network architecture and the corresponding Lipschitz constants.

In this work, we propose an analytical expression for the Lipschitz constant, tailored to different neural network architectures. This analytical approach provides a deeper understanding of how network design influences robustness, particularly in robotic perception systems where maintaining reliability under diverse environmental conditions is crucial. Through experiments, we validate the accuracy of our analytical expression and investigate how network depth, width, and other architectural choices impact robustness. Our goal is to identify architectures that achieve minimal Lipschitz constants, thereby maximizing robustness under comparable accuracy levels — a key consideration for safe and robust robot learning systems deployed in the real world.
Our contributions can be summarized as: 
\begin{itemize}[leftmargin=12pt,itemsep=0pt,topsep=0pt]
    \item We present an expression for the Lipschitz constant based on the architecture of neural networks, providing valuable insights into designing robust neural networks for robot learning systems.
    \item The experimental results validate the proposed mathematical expression and demonstrate the relationship between neural network architecture and its robustness in perception tasks.
    \item In conclusion, our findings indicate that wider networks tend to be more robust than narrower ones, shallower networks exhibit greater robustness compared to deeper ones, and neural networks with weights of lower variance are generally more robust than those with higher variance.
\end{itemize}

\section{Related Works}
\textbf{Robust Learning in Robotics}. Robust robot learning focuses on the development of autonomous systems that can operate reliably and safely within dynamic and uncertain environments \cite{mueller2018robust, abuduweili2021robust}. Various approaches have been implemented to enhance robustness, including safe reinforcement learning, which aims to optimize policies while ensuring safety constraints \cite{brunke2022safe, zhao2024guard}, and techniques like domain adaptation \cite{bousmalis2018using} and online adaptation \cite{abuduweili2020robust, abuduweili2023online}, which facilitate system adaptability across diverse environments. Furthermore, methods such as adversarial training and imposing constraints on the Lipschitz constant have shown promise in enhancing the robustness of robot learning systems by mitigating vulnerability to environmental variations and adversarial inputs \cite{chen2020adversarial, weng2018evaluating}.

\textbf{Adversarial Training and Verification}.  Adversarial training is a commonly used technique for improving the robustness of a neural network  \cite{goodfellow2014explaining}.  However, while such an adversarially trained network is made robust to some attacks in training, it can still be vulnerable to unseen attacks. 
Then neural Network Verification can be used to prove a lower bound on the robustness of neural networks \cite{liu2021algorithms}. 
Certified robust training under verification, focusing on training neural networks with certified and provable robustness – the network is considered robust on an example if and only if the prediction is provably correct for any perturbation in a predefined set \cite{zhang2018efficient,wei2024improve}.

\textbf{Lipschitz constant and Robustness}. Different from certified robust training, some researchers bound the sensitivity of the function to input perturbations by bounding the Lipschitz constant to certify or improve the robustness of neural networks \cite{weng2018evaluating}.  Weng \textit{et al.} \cite{weng2018towards} convert the robustness analysis problem into a local Lipschitz constant estimation problem.  
Designing and training neural networks with bounded Lipschitz constant is a promising way to obtain certifiably robust classifiers \cite{zhang2022rethinking}. Many works handle the Lipschitz bound by
leveraging specific mathematical properties such as the spectral norm \cite{yoshida2017spectral,nguyen2021tight}.  
Unlike previous studies, this work focuses on deriving an analytical expression for the Lipschitz constant based on the architecture of a neural network. We investigate the relationship between neural network architecture and robustness. While prior research typically calculates the Lipschitz constant using exact network parameters and numerical methods, our approach emphasizes the role of architecture rather than specific parameters in estimating the Lipschitz constant.


\section{Lipschitz Continuity of Neural Networks}
\textbf{Notation.} 
We use boldface letters to denote vectors (e.g., $\vec{x}$) or vector functions (e.g., $\vec{f}$), and use $x_i$ or $f_i$ to denote its $i$-th element.  We use capital letters to denote  matrices (e.g., $\mat{W}$), and use $W_{i,j}$ to denote the element of $i$-th row and $j$-th column. For a unary function $\sigma$, $\sigma(\vec{x})$ applies $\sigma(\cdot)$ element-wise on vector $\vec{x}$. The $\l_p$-norm ($p \ge 1$) and $l_\infty$-norm of a vector $\vec{x}$ are defined as $\|\vec{x}\|_p = (\sum_i |x_i|^p)^{\frac{1}{p}}$   and
$\|\vec{x}\|_\infty = \max_i |x_i|$, respectively. 
In the following, we consider a  real-valued function $f: \mathbb{R}^n \rightarrow \mathbb{R}^m$. We mainly consider perturbation in $l_p$ norm, e.g. $|| \delta ||_p$.  

\subsection{Neural Networks}
\label{sec:nn}
In this work, we consider standard neural networks composed of affine layers (such as multilayer perceptrons or convolutional layers) combined with element-wise activation functions. 
\begin{align}
    \vec{x}^{(l)} = \sigma^{(l)}(\vec{\tilde{x}}^{(l)}), \quad \vec{\tilde{x}}^{(l)} = \mat{W}^{(l)} \vec{x}^{(l)-1} + \vec{b}^{(l)} 
\end{align}
Here $M$ is the number of layers and usually $\sigma^{(M)}(\vec{x})=\vec{x}$ is the identity function. The network takes $\vec{x}^{(0)}:=\vec{x}$ as the input and outputs $\vec{x}^{(M)}:=\vec{y}$. We use $\vec{f}$ to denote the whole neural network functions:
\begin{align}
    \vec{f}(\vec{x}) = \sigma^{(l)}(\mat{W}^{(l)} (\cdots  \sigma^{(1)}( \mat{W}^{(1)} \vec{x} + \vec{b}^{(1)} ) )  + \vec{b}^{(l)})
\end{align}

In this work, we mainly consider the classification tasks. 
Let $\vec{x} \in \mathbb{R}^n$ be an input vector of a $m$-class classification function $\vec{f}: \mathbb{R}^n \rightarrow \mathbb{R}^m$. The predicted logits (with softmax activation) is ${\vec{y}}=[{y}_1, {y}_2, \cdots, \hat{y}_m] = \vec{f}(\vec{x})$, and the predicted class is given as $c(\vec{x}) = \arg \max_{1 \le i \le m} {y}_i$. 

\subsection{Lipschitz Continuous} 
\label{sec:lip}
\textbf{Definition 1} (Lipschitz continuous.) A function $\vec{f}: \mathbb{R}^n \rightarrow \mathbb{R}^m$ is called $L$-Lipschitz continuous w.r.t. norm $\|\cdot\|$ if there exists a constant $L$ for any pair of inputs $\vec{x}, \vec{y} \in \mathbb{R}^m$, such that
\begin{align}
\| \vec{f}(\vec{x}) - \vec{f}(\vec{y}) \| \le L \| \vec{x} - \vec{y}  \|.
\end{align}
The smallest $L$ for which the previous inequality is true is called the Lipschitz constant of $\vec{f}$.
For local Lipschitz functions (i.e. functions whose restriction to some neighborhood around any point is Lipschitz), the Lipschitz constant may be computed using its differential operator.

\textbf{Theorem 1} (Rademacher \cite{federer2014geometric}, Theorem 3.1.6)  If  $\vec{f}: \mathbb{R}^n \rightarrow \mathbb{R}^m$ is is a Lipschitz continuous function, and $\vec{f}$ is differentiable almost everywhere, then 
\begin{align}
    L = \sup_{\vec{x}} \| \nabla \vec{f}(\vec{x}) \|,
\end{align}
where $\mat{J}: = \nabla \vec{f}(\vec{x}) = \frac{\partial  \vec{f}}{\partial  \vec{x}}$ is the Jacobian matrix, and $\| \mat{J} \| = \sup_{\vec{u}, \| \vec{u}\|=1} \| \mat{J} \vec{u}\|$  is the operator norm of the  Jacobian matrix. 
According to the Min-max principle for singular values \cite{horn2012matrix}, the largest singular value $s_{max}(\mat{M})$ of a Matrix $\mat{M}$ is equal to the operator norm: $s_{max}(\mat{M}) = \| \mat{M} \|$. Then we have the following proposition.

\textbf{Proposition 1}. (Estimating the Lipschitz constant.) 
Lipschitz constant $L$ of a function $\vec{f}$ can be estimated by the largest singular value of its Jacobian $\mat{J} = \nabla \vec{f}(\vec{x})$:
\begin{align}
   L = \|\mat{J} \| =  s_{max}(\mat{J}).
\end{align}

\subsection{Lipschitz Constant and Robustness} 
\label{sec:lip_robust}
If the neural network $\vec{f}$ has a small Lipschitz constant $L$, then $L$-Lipschitz continuity implies
that the change of network output can be strictly controlled under input perturbations.

\textbf{Definition 2} (Perturbed example and adversarial example.) 
Let $\vec{x} \in \mathbb{R}^n$ be an input vector of a classification function $\vec{f}: \mathbb{R}^n \rightarrow \mathbb{R}^m$, and the predicted class is given as $c(\vec{x}) = \arg \max_{1 \le i \le m} y_i,  y_i = f_i(\vec{x})$. Given $\vec{x}$, we say $\vec{x_a} =\vec{x} + \vec{\delta}$ is a perturbed example of $\vec{x}$ with noise $\vec{\delta} \in \mathbb{R}^n$ under $l_p$ perturbation $\|\vec{\delta} \|_p = \epsilon$.  An adversarial example is a perturbed example $\vec{x_a}$ that changes the predicted class $c(\vec{x})$. 

\textbf{Definition 3} (Margin of a prediction.) The margin of a prediction denotes the difference between the largest and second-largest output logits: 
\begin{align}
    \text{margin}(\vec{f(x)}) = \max(0, y_t - \max_{i \ne t} y_i)
\end{align}
where $\vec{y}=[y_1, y_2, \cdots] = \vec{f}(\vec{x})$ is the predicted logits from the model $\vec{f}$ on data point $\vec{x}$. $y_t$ is the correct logit ($\vec{x}$ belongs to $t$-th class). We assume that the original prediction is correct: i.e. $y_t = \arg \max_i y_i$.  
According to the results from Li \textit{et al.} \cite{li2019preventing}, we have the sufficient condition for a data point to be provably robust to perturbation-based adversarial
examples:

\textbf{Theorem 2} (Li \cite{li2019preventing}) If  $2^{\frac{p-1}{p}}\cdot L \cdot \epsilon < \text{margin}(\vec{f(x)})$, where $\vec{f}$ is a $L$-Lipschitz continuous function under $l_p$ norm, then $\vec{x}$ is robust to any input perturbation $\vec{\delta}$ with $\|\vec{\delta} \|_p \le \epsilon$.

Specifically, we have a proposition for $l_2$ or $l_\infty$ perturbation by simply letting $p=2$ or $p=\infty$ in Theorem 2.

\textbf{Proposition 2}. (Certified Robustness of Lipschitz networks.) For a neural network classifier $\vec{f}: \mathbb{R}^n \rightarrow \mathbb{R}^m$ with Lipschitz constant $L$. Then the neural network classifier is provably robust under $l_2$ perturbation $\|\vec{\delta} \|_2 \le \frac{\sqrt{2}}{2L} \text{margin}(\vec{f(x)})$ or provably robust under $l_\infty$ perturbation $\|\vec{\delta} \|_\infty \le \frac{1}{2L} \text{margin}(\vec{f(x)})$.

\subsection{Maximum Singular Value of Random Matrices} 
\label{sec:rmt}
As demonstrated in proposition 2,  the Lipschitz constant determines the certified robustness of a neural network classifier unde $l_p$ norm perturbation. Our goal is to derive an analytical expression for the Lipschitz constant based on a given neural network architecture and explore the relationship between the architecture and robustness.
According to \ref{sec:lip}, the Lipschitz constant corresponds to the largest singular value of the neural network's Jacobian matrix. Thus, the challenge is to estimate this largest singular value. In this study, we propose an approximation method using Random Matrix Theory (RMT) \cite{akemann2011oxford} to estimate the Lipschitz constant. Although this method does not yield the exact value of the Lipschitz constant, the approximation remains valuable for exploring its analytical relationship with neural network architectures.



\textbf{Theorem 3} (Rudelson \cite{rudelson2010non}). 
Let $\mat{A}$ be an $N \times n$ random matrix whose entries are independent copies of some random variable with zero mean, unit variance, and finite fourth moment. Suppose that the dimensions $N$ and $n$ grow to infinity while the aspect ratio $n/N$ converges to
some number $y \in (0,1]$, $n/N \rightarrow y$. Then the maximum singular value $s_{max}(\mat{A})$ converges to: 
\begin{align}
     \frac{1}{\sqrt{N}} s_{max}(\mat{A}) \rightarrow  1 + \sqrt{\frac{n}{N}} \quad \text{almost surely}. \label{eq:s_max}
\end{align}
Thus, asymptotically, the expectation of the maximum singular value is $\mathbb{E} s_{max}(\mat{A}) \approx \sqrt{N} + \sqrt{n}$. 
For the maximum singular value of the product of matrix $\mat{A}$ and a scaler $\alpha$, we have:
\begin{align}
  s_{max}(\alpha \mat{A}) = \| \alpha \mat{A} \|  =  \alpha \| \mat{A} \| = \alpha  s_{max}(\mat{A}).
\end{align}

\Cref{eq:s_max} provides the maximum singular value for matrices with unit variance. If the variance is  $\alpha^2$, it can be approximated by scaling the maximum singular value of the unit variance matrix by $\alpha$. Thus, we propose the following expression for the maximum singular value of random matrices with variance $\alpha^2$.

\textbf{Proposition 3}. (Maximum singular value of a random matrix.) Let $\mat{A}$ be an $N \times n$ random matrix whose entries are independent copies of some random variable with zero means, $\alpha^2$ variance, and finite fourth moment. The expectation of the maximum singular value can be approximated by $\mathbb{E} s_{max}(\mat{A}) \approx \alpha ( \sqrt{N} + \sqrt{n})$.   

\subsection{Variance of the Jacobian}
Proposition 3 provides an approximation method for estimating the singular value of a random matrix.  The Jacobian matrix of a neural network is random at initialization due to the randomly initialized network parameters. Therefore, at least during initialization, we can estimate the Lipschitz constant of a neural network using Propositions 3 and 1. Furthermore, even during training, the network parameters remain nearly random for wide neural networks, based on neural tangent kernel analysis \cite{jacot2018neural}. Thus, in this subsection, we estimate certain statistical properties (e.g., mean and variance) of the Jacobian matrix, which can be used to approximate the Lipschitz constant.

\begin{figure}[htbp]
\centering
\includegraphics[width=0.5\linewidth]{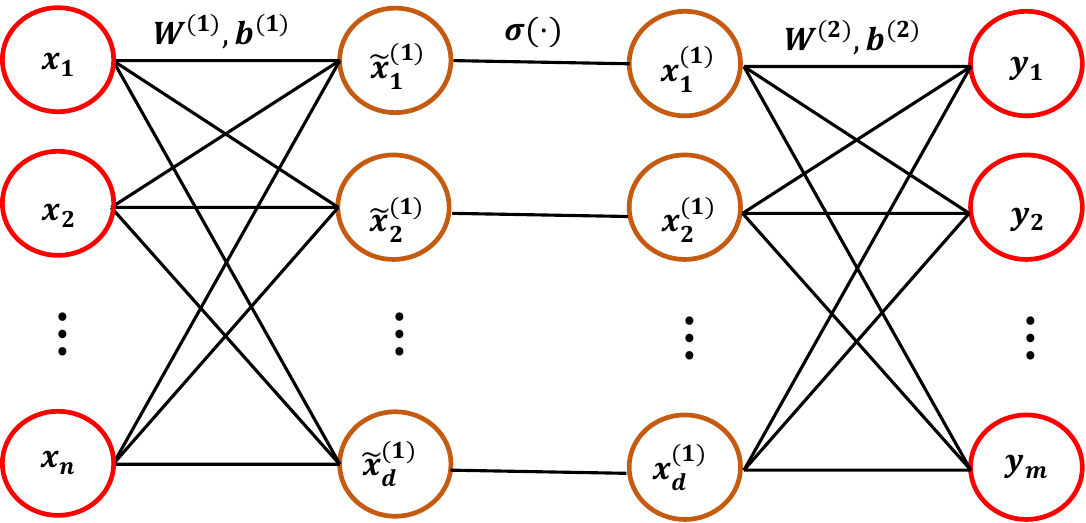}
\vspace{-2mm}
\caption{Neural Network Architectures.}
\label{fig:nn}
\end{figure}

As an illustration, we first consider a multilayer perceptron (MLP) with one hidden layer, as shown in \cref{fig:nn}. The analysis can be extended to MLPs with additional hidden layers due to the application of the chain rule for derivatives.  
In \cref{fig:nn}, $n$ represents the input dimension, $m$ is the output dimension, and $d$ is the hidden dimension. The network parameters include weights $\mat{W}^{(1)}, \mat{W}^{(2)}$, and biases $\vec{b}^{(1)}, \vec{b}^{(2)}$.  The activation function is denoted as
$\sigma(\cdot)$. We initialize these parameters using Xavier initialization \cite{glorot2010understanding}: 
\begin{align}
{W}^{(1)}_{i,j} \sim \mathbb{N}(0, \frac{2 \alpha^2}{d + n}), \quad  {b}^{(1)}_{i}  = 0, ~
{W}^{(2)}_{i,j} \sim \mathbb{N}(0, \frac{2 \alpha^2}{m + d}), \quad  {b}^{(2)}_{i}  = 0,
\end{align}
We can compute the expectation and variance of the Jacobian as follows. Please refer to \cref{sec:ap_var_jac} for further details.
\begin{align}
    \mathbb{E}[J_{i,j}] &=  \sum_{k =1}^{d}  \mathbb{E}[W^{(2)}_{i,k}]  \mathbb{E}[\sigma'(\tilde{x}_k^{(1)})] \mathbb{E}[ W^{(1)}_{k,j}] = 0 \\
    \mathbb{VAR}[J_{i,j}] &=  \sum_{k =1}^{d}  \mathbb{VAR}[W^{(2)}_{i,k}]  \mathbb{VAR}[\sigma'(\tilde{x}_k^{(1)})] \mathbb{VAR}[ W^{(1)}_{k,j}]  =  \frac{4d}{(d+n)(d+m)} \alpha^4 q^2 
\end{align}
where $q^2 = \mathbb{VAR}[\sigma'(x)]$ denotes the variance of random variables $\sigma'(x)$, when $x \sim \mathbb{N}(0,1)$.For example, with the ReLU activation function, $q^2 = \mathbb{VAR}[\sigma'(x)]=\frac{1}{4}$ when $x$ follows a standard normal distribution.
By applying the chain rule of derivatives,  this approach can be extended to compute the variance of the Jacobian matrix for any $M$-layer network, leading to the following proposition.

\textbf{Proposition 4}. (Expectation and Variance of a Jacobian matrix.) Let $\vec{f}$ denote the $M$-layer neural network with input dimension $n$, output dimension $m$, and hidden dimension $d$. The Jacobian matrix is denoted as $\mat{J} = \frac{\partial \vec{f(x)}}{\vec{x}}$.Assume the neural network is initialized using Xavier initialization. Then, the expectation and variance of the Jacobian can be estimated as:
\begin{align}
    \mathbb{E}[J_{i,j}] =0, \quad \mathbb{VAR}[J_{i,j}] = \frac{4d}{(d+n)(d+m)} \alpha^{2M} q^{2M-2},   
\end{align}
where $q^2 = \mathbb{VAR}[\sigma'(x)]$ denotes the variance of $\sigma'(x)$ for $x \sim \mathbb{N}(0,1)$. For ReLU, $q=\frac{1}{2}$.

\subsection{Lipschitz Constant and Robustness of Neural
Networks}
\label{sec:lip_robust_nn}
By combining Proposition 4, Proposition 3, and Proposition 1, we derive an analytical expression for the Lipschitz constant of neural networks

\textbf{Proposition 5}. (Expectation of  Lipschitz Constant for neural networks.) Let $\vec{f}: \mathbb{R}^n \times  \mathbb{R}^m$ denote the $M$-layer feedforward neural network with input dimension $n$, output dimension $m$, and hidden dimension $d$.  Let $\sigma(\cdot)$ denote the activation function.
Assume the neural network was initialized with Xavier initialization, such that the weight matrix $W^{(l)} \in \mathbb{R}^{n^{(l)} \times n^{(l-1)}}$ satisfies the statistical properties $\mathbb{E}[W^{(l)}] = 0, \mathbb{VAR}[W^{(l)}] = \frac{2 \alpha^2}{n^{(l)}+ n^{(l-1)}}$. Then, the expectation of the Lipschitz constant can be estimated as:
\begin{align}
    \mathbb{E}[L] = \frac{2 \sqrt{d}}{\sqrt{(d+n)(d+m)}} \alpha^{M} q^{M-1} (\sqrt{n} + \sqrt{m}), \label{eq:lip}
\end{align}
where $q^2 = \mathbb{VAR}[\sigma'(x)]$ denotes the variance of $\sigma'(x)=\frac{\partial \sigma(x)}{\partial x}$ for $x \sim \mathbb{N}(0,1)$. For ReLU, $q=\frac{1}{2}$.

Proposition 2 illustrates the relationship between robustness and the Lipschitz constant. Using the Lipschitz constant as a bridge between robustness and neural network architecture, and combining Proposition 2 with Proposition 5, we derive the following properties:
\begin{itemize}[leftmargin=12pt,itemsep=0pt,topsep=0pt]
    \item If $\alpha q >1$, increasing the number of network layers (depth $M$) results in an increase in the Lipschitz constant, leading to decreased robustness.
    \item  Increasing the number of hidden dimensions (width $d$) decreases the Lipschitz constant, thereby increasing robustness.
    \item Increasing the weight variance $\alpha$ results in an increase in the Lipschitz constant, which in turn decreases the robustness of the neural network.
\end{itemize}

The exact values of the neural network weights change during training, meaning the analytical estimation in \cref{eq:lip} may not be accurate after training. The critical point is whether the distribution of the Jacobian matrix, as described in \cref{eq:J}, satisfies the conditions of Theorem 3 (singular value of random matrices). Notably, Theorem 3 does not require the matrix's entries to follow a Gaussian distribution; it generally requires that the distribution has a zero third moment and a finite fourth moment. With some modifications, such as normalization $(J_{i,j} - \mu)/\sigma$, the distribution of the Jacobian's elements may still meet these conditions.
At the very least, this condition holds true for sufficiently wide neural networks. The Neural Tangent Kernel (NTK) Theory suggests that, in infinitely wide networks, the weights remain nearly constant during training  \cite{jacot2018neural}.   

In this work, we assume that the Jacobian after training continues to satisfy the zero third-moment and finite fourth-moment conditions. We defer a detailed analysis and evaluation of this assumption to future work. Under this assumption, the Lipschitz constant of a neural network after training can still be estimated using \cref{eq:lip}. The primary difference from the initialization phase is the use of the variance after training, $\tilde{\alpha}$,  to replace the initial variance $\alpha$  of weights.  Specifically, for a weight matrix $W^{(l)} \in \mathbb{R}^{n^{(l)} \times n^{(l-1)}}$, we have  $\mathbb{VAR}[W^{(l)}_{i,k}(t=0)] = \frac{2 \alpha^2}{ n^{(l)}+ n^{(l-1)}}$ at initialization. After training for $t^*$ steps, the variance becomes $\mathbb{VAR}[W^{(l)}_{i,k}(t=t^*)] = \frac{2 \tilde{\alpha}^2}{ n^{(l)}+ n^{(l-1)}}$.  By measuring $\tilde{\alpha}$,we can then estimate the Lipschitz constant using \cref{eq:lip}.

\section{Experiments}
\subsection{Evaluating the correctness of Proposition 5}
One of the contributions of this work is the proposal of an analytical expression for the Lipschitz constant of neural networks, as given in \cref{eq:lip}.In this section, we evaluate the accuracy of this analytical expression through toy experiments. We design various neural network architectures with different numbers of layers $M$, hidden dimensions $d$, and weight variances $\alpha^2$. At initialization, we estimate the Lipschitz constant and compare the analytical estimation from \cref{eq:lip} with numerical measurements obtained using bound propagation \cite{shi2022efficiently}. The numerical method provides an accurate approximation of the Lipschitz constant.
To minimize the impact of randomness, we conduct 10 trials for each neural network configuration and report the average results. Additionally, we apply moving average smoothing to filter the curves for better clarity.

\begin{figure}[htbp]
\vspace{-2mm}
\centering
\subfloat[$L$ v.s. depth $M$]{
\includegraphics[width=0.33\linewidth]{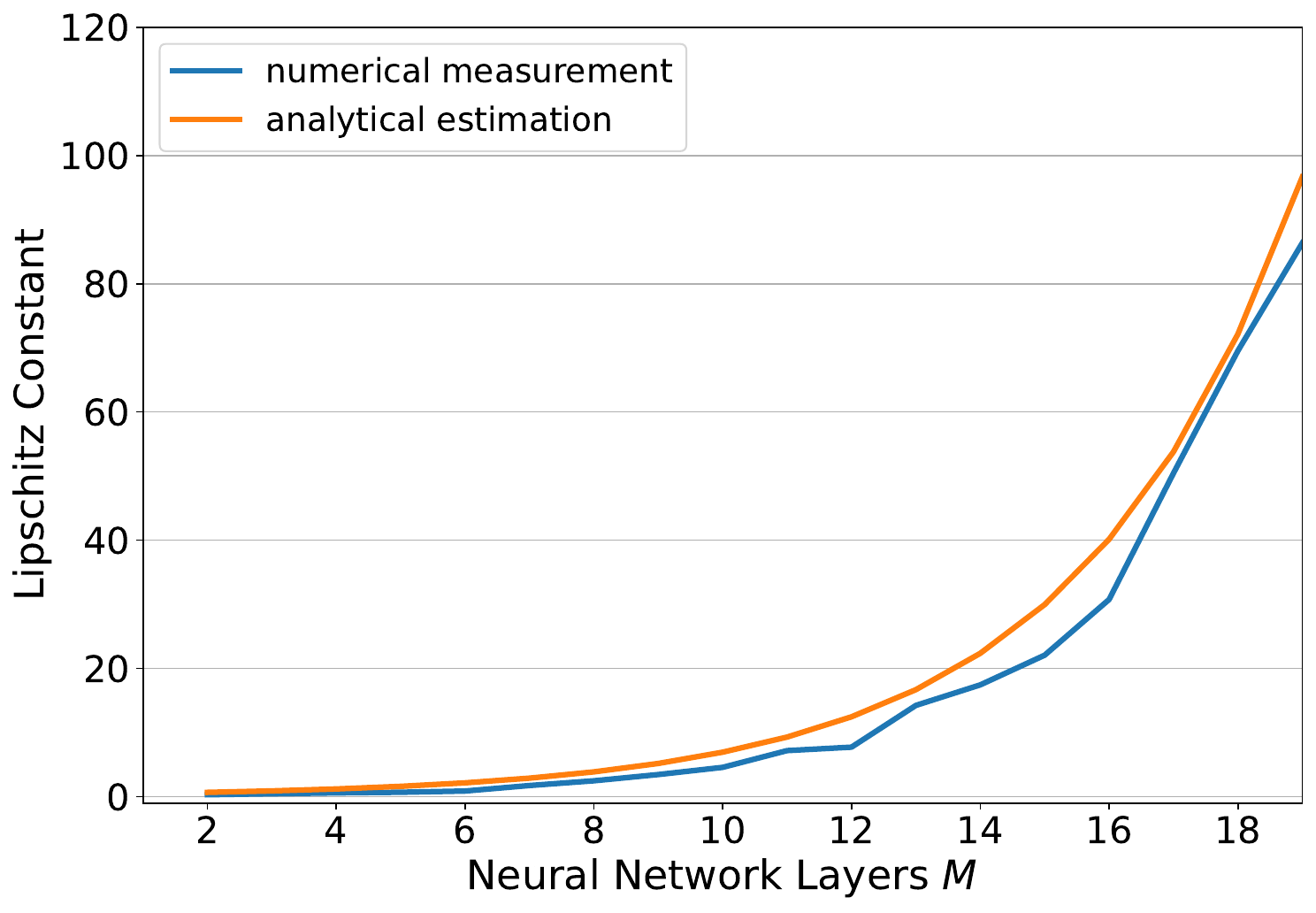}
\label{fig:lip_depth}}   
\subfloat[$L$ v.s. hidden dimension $d$ ] {
\includegraphics[width=0.33\linewidth]{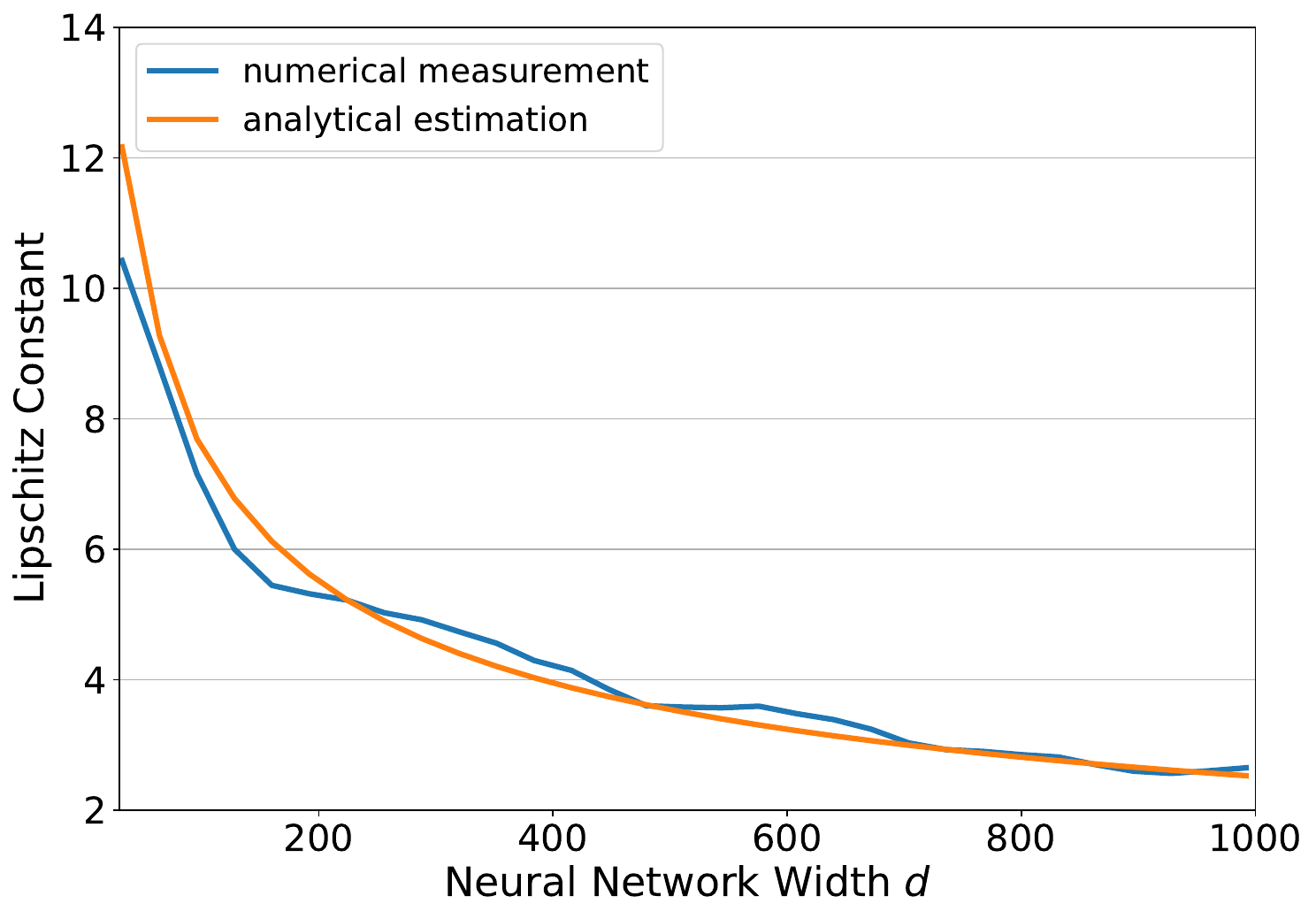}
\label{fig:lip_width}}  
\subfloat[$L$ v.s. weight variance $\alpha$] {
\includegraphics[width=0.33\linewidth]{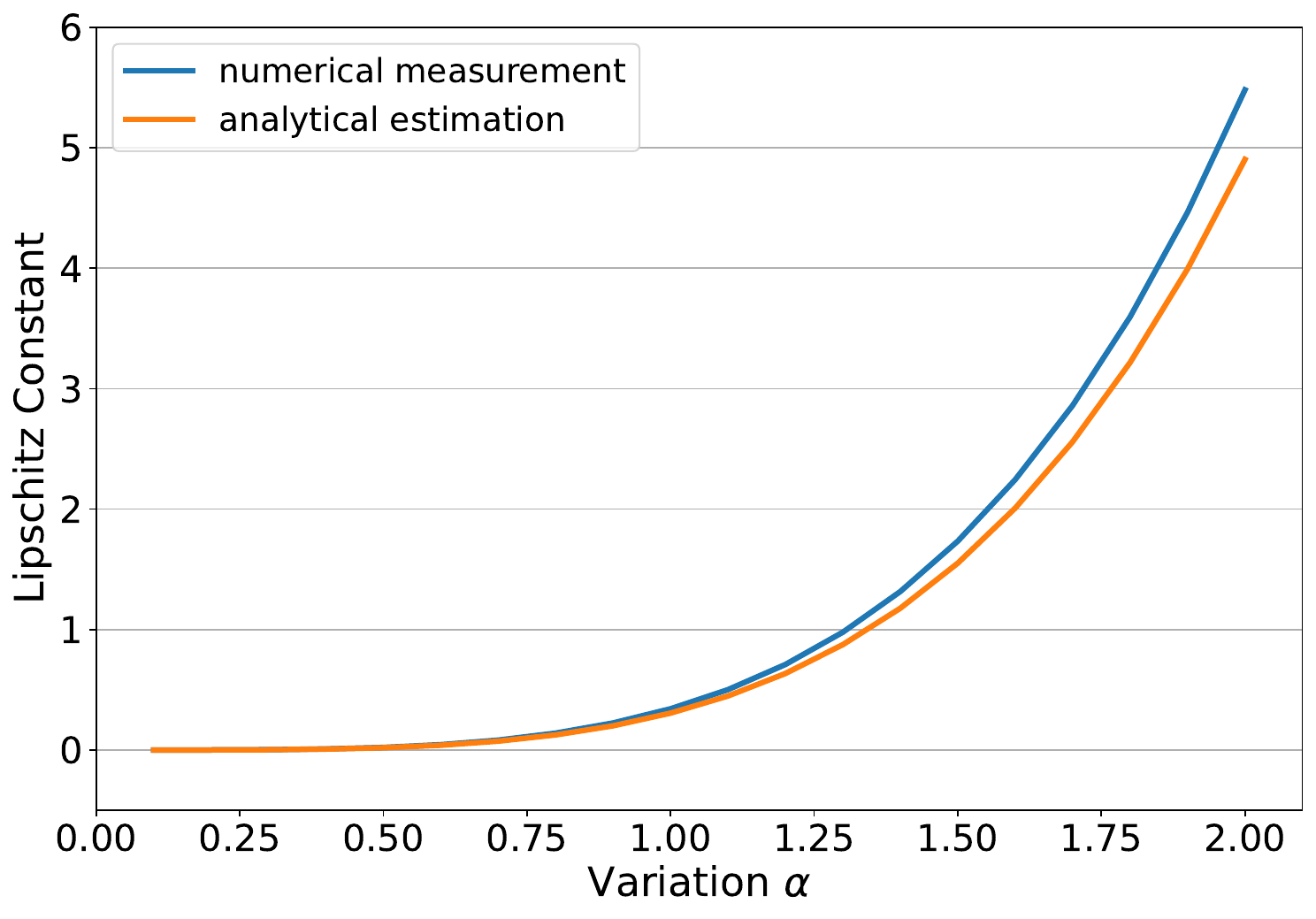}
\label{fig:lip_sigma}}
\caption{Comparison of estimated Lipschitz constant between analytical estimation \cref{eq:lip} and numerical measurement. }
\label{fig:lip_cmp}
\vspace{-2mm}
\end{figure}

\Cref{fig:lip_cmp} presents the experimental results comparing the Lipschitz constant estimates from the proposed analytical expression in \cref{eq:lip} and the numerical measurements from previous work  \cite{shi2022efficiently}. \Cref{fig:lip_depth} illustrates the Lipschitz constant behavior across different numbers of layers (depth) in neural networks. In this experiment, the width of all networks is set to $d=256$ and $\alpha=2$. As shown, the proposed analytical estimation closely matches the numerical measurement. In summary, as the network depth ($M$) increases, the Lipschitz constant increases exponentially, following the trend $L_0^M$. \Cref{fig:lip_width} shows the Lipschitz constant as a function of network width. For this experiment, we use 4-layer networks with $\alpha=2$. Again, the analytical estimation aligns closely with the numerical measurement. The results indicate that as the number of hidden dimensions (width $d$) increases, the Lipschitz constant decreases, following the trend $L_0/\sqrt{d}$. Finally, \cref{fig:lip_sigma} demonstrates the Lipschitz constant behavior under different weight variances $\alpha$. The proposed analytical estimation continues to match the numerical measurement. These results validate the correctness of the proposed analytical estimation of the Lipschitz constant for multilayer perceptrons.

\subsection{Simple perception experiments}
The analytical estimation of the Lipschitz constant from \cref{eq:lip} may not be entirely accurate after training. However, it still offers valuable insights into trained networks, particularly the relationship between neural network architecture and robustness. In this section, we conduct a visual perception experiment using the MNIST image classification dataset \cite{lecun1998mnist}. In the experiment, we design various neural architectures with different numbers of layers $M$, hidden dimensions $d$, and weight variances $\alpha^2$. We train the model using Stochastic Gradient Descent until converges. We train the models using Stochastic Gradient Descent (SGD) until convergence. After training, we evaluate the test set accuracy and compute the certified accuracy using a verification method. Specifically, for each image, we assess the robustness of the neural network under an $\| \delta \|_2 \le 1$ perturbation using Interval Bound Propagation (IBP) \cite{gowal2018effectiveness}. We then calculate the average certified accuracy under these perturbations.

\begin{table}[h]
\caption{ Standard accuracy, certified accuracy, Lipschitz constant of different network architectures. }
\label{tab:result}\centering
\begin{tabular}{cc|cc|c}
\hline
Layer & Width & Accuracy & Certified Accuracy & Lipschitz Constant \\ \hline
2 & 128 & 0.942 & 0.864 & 76.0 \\
2 & 256 & 0.943 & 0.866 & 73.2 \\
2 & 512 & 0.944 & 0.871 & 68.9 \\ \hline
3 & 128 & 0.956 & 0.806 & 89.5 \\
3 & 256 & 0.957 & 0.814 & 81.3 \\
3 & 512 & 0.961 & 0.816 & 81.1 \\ \hline
4 & 128 & 0.965 & 0.582 & 102.5 \\
4 & 256 & 0.968 & 0.582 & 98.1 \\
4 & 512 & 0.971 & 0.593 & 94.5 \\ \hline
\end{tabular}
\vspace{-2mm}
\end{table}

\begin{figure}[htbp]
\vspace{-2mm}
\centering
\subfloat[Lipschitz constant v.s. depth $M$ ]{
\includegraphics[width=0.4\linewidth]{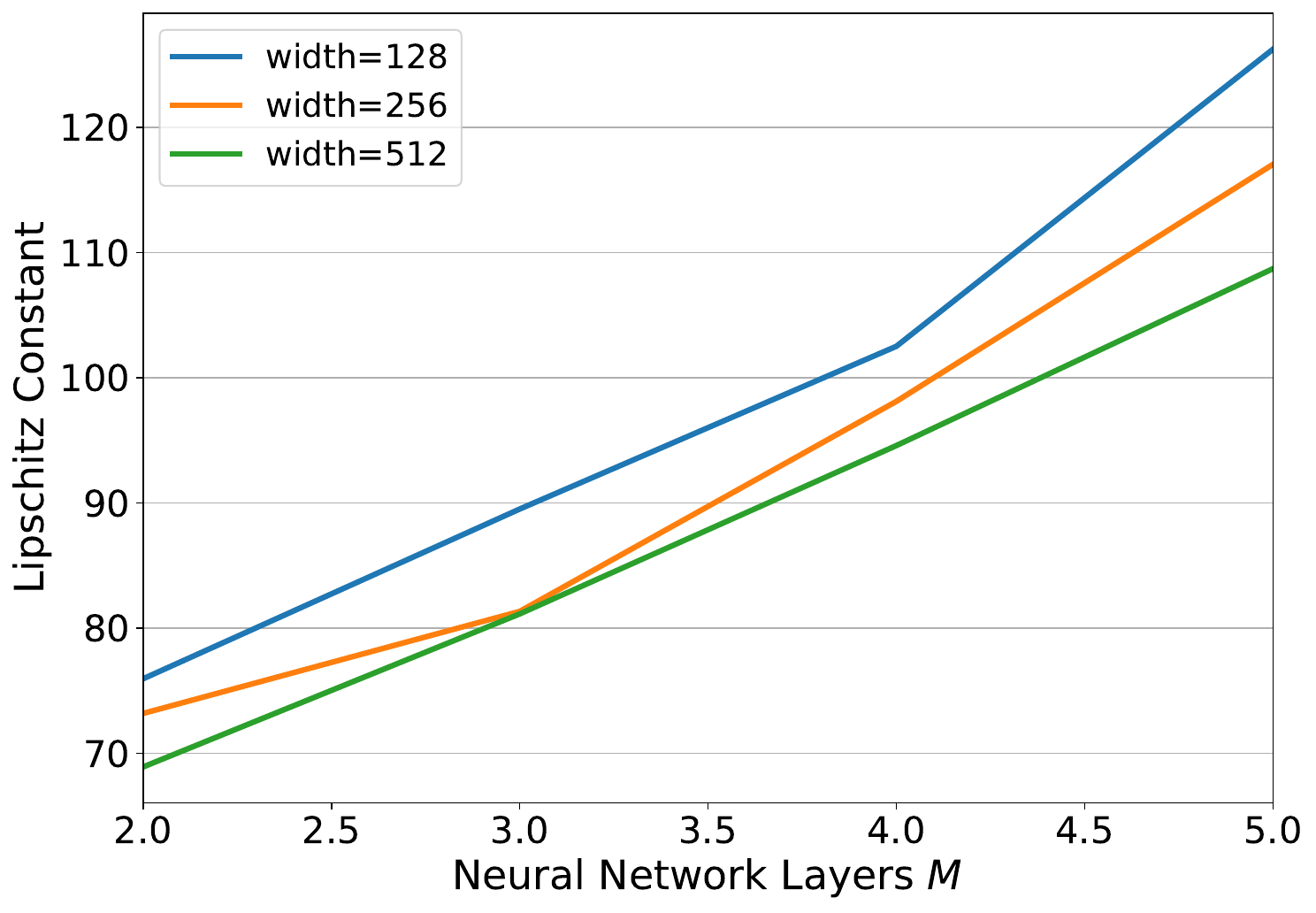}}   \vspace{-1mm}
\subfloat[Certified accuracy v.s. depth $M$ ] {
\includegraphics[width=0.4\linewidth]{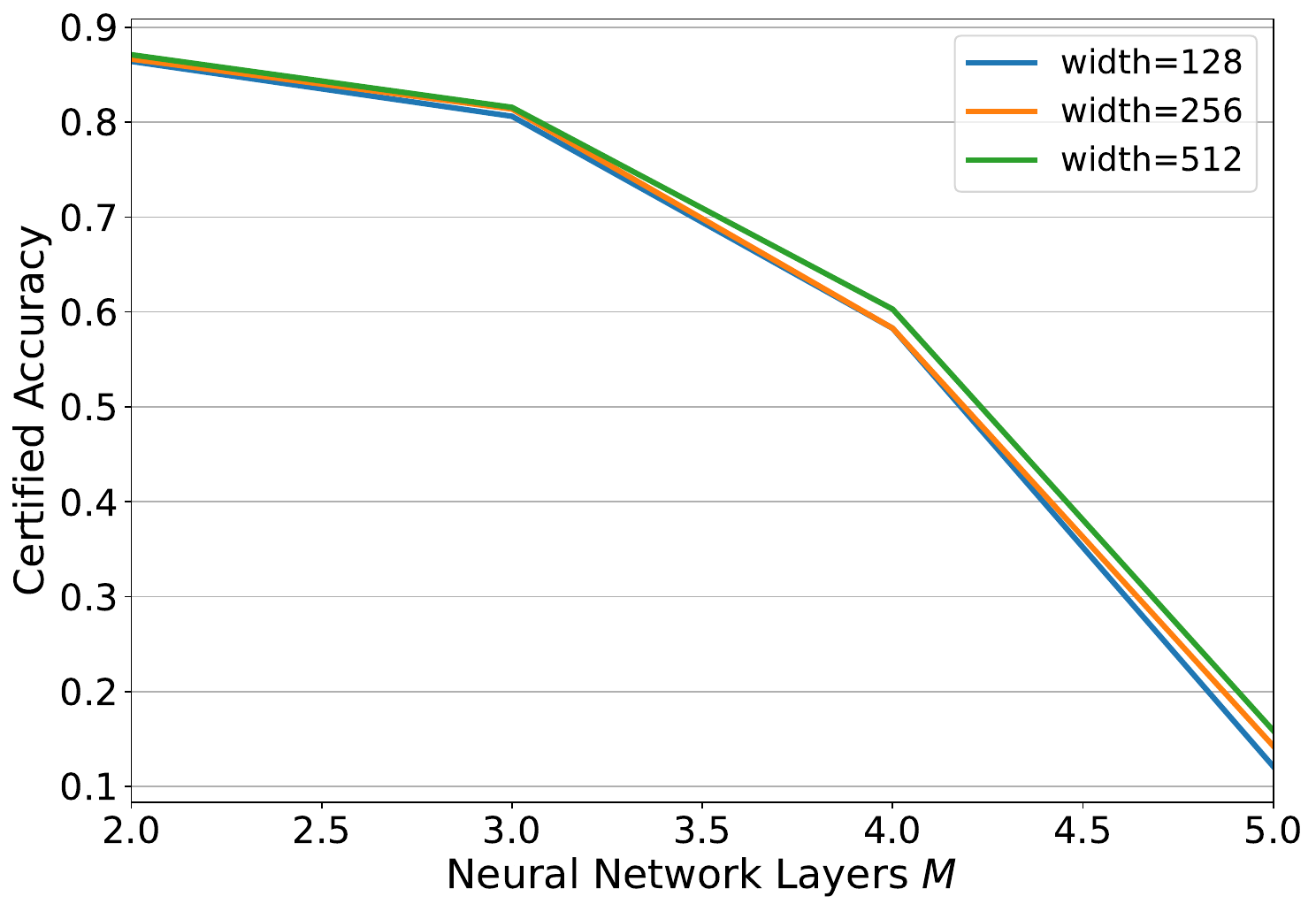}} 
\vspace{-1mm}
\caption{ Lipschitz constant and Certified accuracy of different neural network architectures.}
\label{fig:mnist_lip}
\vspace{-2mm}
\end{figure}

The experimental results are presented in \cref{tab:result}. While all network architectures achieve similar standard accuracy, their certified accuracy and Lipschitz constant vary significantly. These results align with the analytical conclusions discussed in \cref{sec:lip_robust_nn}. 1) As the number of network layers (depth $M$) increases, the Lipschitz constant increases, resulting in decreased robustness.  2) As the number of hidden dimensions (width $d$) increases, the Lipschitz constant decreases, leading to increased robustness. \Cref{fig:mnist_lip} visually illustrates these conclusions for clarity.

We also conduct experiments to examine the impact of different variances $\alpha$. \Cref{tab:var_lip} shows the standard accuracy, certified accuracy, and Lipschitz constant for a 4-layer neural network with a width of $d=128$ under different initial weight variances $\alpha$. In the table, accuracy, certified accuracy, and the Lipschitz constant are measured after training.  $\alpha$ represents the standard deviation of the weights at initialization. While $\alpha$ may change during training, the initial value provides insight into the Lipschitz constant: a larger initial $\alpha$  results in a larger Lipschitz constant and, consequently, reduced robustness. The evolution of the Lipschitz constant and certified accuracy during training is detailed in \cref{sec:ap}. 

\begin{table}[h]
\caption{ Standard accuracy, certified accuracy, Lipschitz constant under different weight variance. }
\label{tab:var_lip}\centering
\begin{tabular}{c|cc|c}
\hline
$\alpha$ & Accuracy & Certified Accuracy & Lipschitz Constant \\ \hline
0.3 & 0.928 & 0.720 & 99.7 \\
1.0 & 0.965 & 0.583 & 102.5 \\
3.0 & 0.974 & 0 & 819.6 \\ \hline
\end{tabular}
\vspace{-2mm}
\end{table}

\section{Conclusions}
Ensuring the robustness of neural networks is essential for the safe and reliable operation of robot learning systems in real-world environments. In this work, we investigated the relationship between neural network architecture and robustness.  We proposed an analytical expression for estimating the Lipschitz constant of neural networks. This expression allows us to evaluate the Lipschitz constant and investigate its connection to network architecture and robustness. Our analytical expression and experimental results suggest that shallower, wider networks tend to exhibit greater robustness.

\textbf{Limitations and Future work.} 
1) This study focuses solely on multilayer perceptrons (MLPs). In future work, we plan to extend our analysis to other network architectures, such as convolutional neural networks (CNNs) and networks with residual connections.
2) The experiments were conducted on the MNIST image classification task. In future research, we aim to test our methods on more complex perception tasks and other robot-learning applications
3) Our current analysis primarily centers on the Lipschitz constant as a key metric for evaluating robustness. However, as noted in Proposition 2, robustness is also affected by the margin of predictions (i.e., the confidence level of the model’s outputs). In future studies, we intend to incorporate the margin of predictions into our robustness analysis and expand our experiments to more complex datasets.

\clearpage


\bibliography{example}  

\newpage
\appendix

\section{More details about the Variance of the Jacobian} 
\label{sec:ap_var_jac}

As an illustration, we first consider a multilayer perceptron (MLP) with one hidden layer, as shown in \cref{fig:nn}. The analysis can be extended to MLPs with additional hidden layers due to the application of the chain rule for derivatives.  
In \cref{fig:nn}, $n$ represents the input dimension, $m$ is the output dimension, and $d$ is the hidden dimension. The network parameters include weights $\mat{W}^{(1)}, \mat{W}^{(2)}$, and biases $\vec{b}^{(1)}, \vec{b}^{(2)}$.  The activation function is denoted as
$\sigma(\cdot)$. We initialize these parameters using Xavier initialization \cite{glorot2010understanding}: 
\begin{align}
{W}^{(1)}_{i,j} &\sim \mathbb{N}(0, \frac{2 \alpha^2}{d + n}), \quad  {b}^{(1)}_{i}  = 0, \\
{W}^{(2)}_{i,j} &\sim \mathbb{N}(0, \frac{2 \alpha^2}{m + d}), \quad  {b}^{(2)}_{i}  = 0,
\end{align}
where $\alpha$ controls the variance of initialization. The scaling of each weight is determined by its corresponding dimensions. The $i$-th coordinate of the final output is:
\begin{align}
    y_i &= \sum_{k =1}^{d} W^{(2)}_{i,k} x_k^{(1)} +  b_i^{(2)} \\
    x_k^{(1)} &= \sigma(\tilde{x}_k^{(1)}), \quad \tilde{x}_k^{(1)} = \sum_{j =1}^{n} W^{(1)}_{k,j} x_j +  b_k^{(1)}
\end{align}
Then the $(i,j)$-th elements of the Jacobian is $\mat{J}=\frac{\partial \vec{y} }{ \partial \vec{x}}$:
\begin{align}
    J_{i,j} &= \frac{\partial y_i }{ \partial x_j} =
     \sum_{k =1}^{d} \frac{\partial y_i }{ \partial x_k^{(1)}} \frac{\partial x_k^{(1)} }{ \partial \tilde{x}_k^{(1)}}  \frac{\partial \tilde{x}_k^{(1)} }{ \partial x_j}  \\
     &=  \sum_{k =1}^{d}  W^{(2)}_{i,k}  \sigma'(\tilde{x}_k^{(1)}) W^{(1)}_{k,j} \label{eq:J}
\end{align}
where $ \sigma'(\tilde{x}_k^{(1)}):=\frac{\partial  \sigma(\tilde{x}_k^{(1)}) }{ \partial \tilde{x}_k^{(1)}}$. Note that, $W^{(2)}_{i,k} $, $\sigma'(\tilde{x}_k^{(1)})$, and $W^{(1)}_{k,j}$ are independent. At initialization, we have:
\begin{align}
     \mathbb{E}[W^{(2)}_{i,k}] &=  \mathbb{E}[W^{(1)}_{i,k}] = 0 \\
     \mathbb{VAR}[W^{(2)}_{i,k}] &= \frac{2 \alpha^2}{d + n}, \quad \mathbb{VAR}[W^{(1)}_{k,j}] = \frac{2 \alpha^2}{d+ m}
\end{align}
Next, we compute the expectation and variance of the Jacobian:
\begin{align}
    \mathbb{E}[J_{i,j}] &=  \sum_{k =1}^{d}  \mathbb{E}[W^{(2)}_{i,k}]  \mathbb{E}[\sigma'(\tilde{x}_k^{(1)})] \mathbb{E}[ W^{(1)}_{k,j}] = 0 \\
    \mathbb{VAR}[J_{i,j}] &=  \sum_{k =1}^{d}  \mathbb{VAR}[W^{(2)}_{i,k}]  \mathbb{VAR}[\sigma'(\tilde{x}_k^{(1)})] \mathbb{VAR}[ W^{(1)}_{k,j}] \\
    &= \frac{2 \alpha^2}{d + n} \cdot \frac{2 \alpha^2}{d + m} \cdot \mathbb{VAR}[\sigma'(\tilde{x}_k^{(1)})] \\
    &=  \frac{4d}{(d+n)(d+m)} \alpha^4 q^2 
\end{align}
where $q^2 = \mathbb{VAR}[\sigma'(x)]$ denotes the variance of random variables $\sigma'(x)$, when $x \sim \mathbb{N}(0,1)$. For example, with the ReLU activation function:
\begin{align}
    \sigma(x) = \max(0,x), \sigma'(x)= \begin{cases}
    1,& \text{if } x > 0\\
    0,              & \text{otherwise}
\end{cases}
\end{align}
Thus, $q^2 = \mathbb{VAR}[\sigma'(x)]=\frac{1}{4}$ when $x$ follows a standard normal distribution.

\section{Additional Experiments}
\subsection{Lipschitz Constant during Training}
\label{sec:ap}

\begin{figure}[htbp]
\vspace{-2mm}
\centering
\subfloat[2-layer neural network ]{
\includegraphics[width=0.8\linewidth]{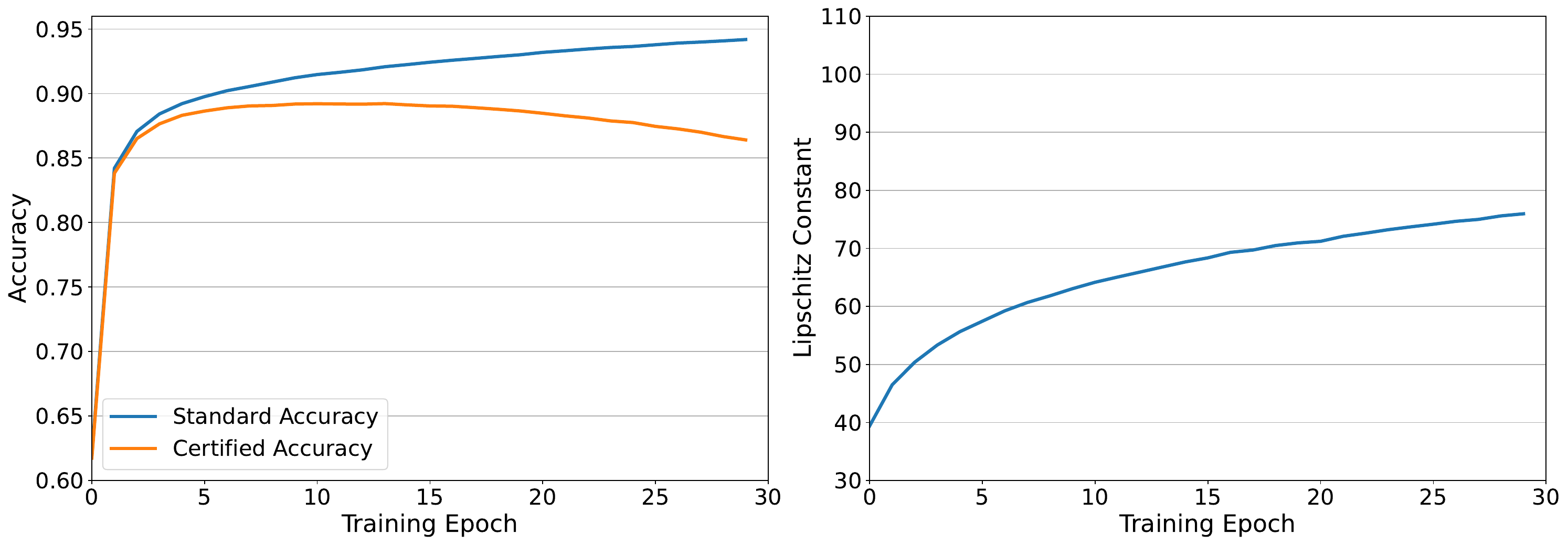}}  \\ \vspace{-1mm}
\subfloat[4-layer neural network ] {
\includegraphics[width=0.8\linewidth]{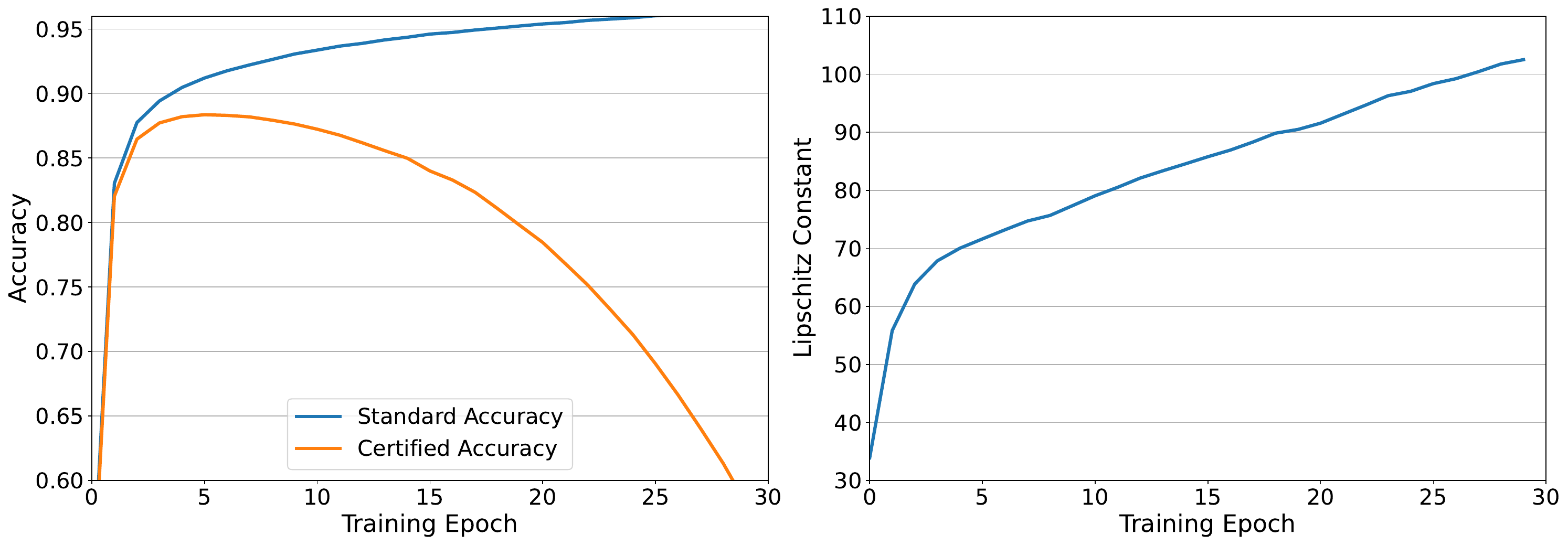}} 
\vspace{-1mm}
\caption{ Lipschitz constant and accuracy during training .}
\label{fig:train_lip}
\vspace{-2mm}
\end{figure}

\Cref{fig:train_lip} shows the standard accuracy, certified accuracy, and Lipschitz constant during training for a 2-layer neural network and a 4-layer neural network. As observed, the Lipschitz constant increases over the course of training. The certified accuracy initially increases in the early stages of training but decreases in the later stages. This decrease in certified accuracy may be attributed to the increase in the Lipschitz constant. Furthermore, \cref{fig:train_lip} indicates that the 2-layer network maintains a smaller Lipschitz constant and higher certified accuracy compared to the 4-layer network.

\subsection{Weight Decay Increases the Robustness of Training}
Weight decay is a common regularization technique used in neural network training.  Let $\vec{f}(\mat{W};\vec{x})$  represent the output of the neural network for an input $\vec{x}$ given the learnable parameter $\mat{W}$. 
Assume the task-dependent objective function (e.g. cross-entropy) is $\mathcal{L}(f(\mat{W};\vec{x}), \vec{y}^\star)$, where $ \vec{y}^\star$ is a ground-truth labels. Then the training objective with weight decay can be formulated as:
\begin{align}
    \min_{\mat{W}}  \mathcal{L}(f(\mat{W};\vec{x}), \vec{y}^\star) + \lambda \| \mat{W} \|, \label{eq:wd}
\end{align}
where $\lambda$ is a penalty term that controls the weight decay. 

We demonstrate that weight decay can help bound the Lipschitz constant during training. For simplicity, we consider a single-layer neural network with either ReLU or Sigmoid activation functions. The network output can be expressed as:
$\vec{f(\mat{W};\vec{x})}=\sigma(\mat{W} \vec{x} + \vec{b})$ , we have:
\begin{align}
    \| \vec{f(\mat{W},\vec{x_1})} - \vec{f(\mat{W},\vec{x_1})} \| &=  \| \sigma(\mat{W} \vec{x_1} + \vec{b}) - \sigma(\mat{W} \vec{x_2} + \vec{b})\| \\
    & \le  \| \mat{W} \vec{x_1} + \vec{b} - \mat{W} \vec{x_2} + \vec{b}\|  = \| \mat{W} \vec{x_1} - \mat{W} \vec{x_2} \| \\
    & \le  \| \mat{W} \| \|  \vec{x_1} - \vec{x_2} \|
\end{align}
The Lipschitz constant $L$ denotes the smallest value for: $\| \vec{f(x_1)} - \vec{f(x_2)} \| \le L \|  \vec{x_1} - \vec{x_2} \|$. We have $L \le \| \mat{W} \|$. For $M$-layer's network, we have a similar property:
\begin{align}
    L \le  \prod_{l=1}^M \| \mat{W^{(l)}} \|
\end{align}
Thus, weight decay serves to minimize the upper bound of the Lipschitz constant. \Cref{tab:wd_lip} presents the experimental results for certified accuracy and the Lipschitz constant under different weight decay values ( varying $\lambda$ in \cref{eq:wd}). As shown, a larger penalty term $\lambda$ results in a smaller Lipschitz constant and, consequently, higher robustness.

\begin{table}[h]
\caption{ Standard accuracy, certified accuracy, Lipschitz constant with weight decay. }
\label{tab:wd_lip}\centering
\begin{tabular}{c|cc|c}
\hline
$\lambda$ & Accuracy & Certified Accuracy & Lipschitz Constant  \\ \hline
0 & 0.965 & 0.583 & 102.6 \\
0.0001 & 0.964 & 0.594 & 100.9 \\
0.001 & 0.962 & 0.675 & 87.0 \\
0.01 & 0.940 & 0.841 & 40.1 \\ \hline
\end{tabular}
\vspace{-2mm}
\end{table}

\end{document}